# Accurate reconstruction of image stimuli from human fMRI based on the decoding model with capsule network architecture

Kai Qiao, Chi Zhang, Linyuan Wang, Bin Yan, Jian Chen, Lei Zeng, Li Tong*
National Digital Switching System Engineering and Technological Research Center, Zhengzhou, China, 450001

*Correspondence author:
Li Tong
tttocean@163.com

Abstract

In neuroscience, all kinds of computation models were designed to answer the open question of how sensory stimuli are encoded by neurons and conversely, how sensory stimuli can be decoded from neuronal activities. Especially, functional Magnetic Resonance Imaging (fMRI) studies have made many great achievements with the rapid development of the deep network computation. However, comparing with the goal of decoding orientation, position and object category from activities in visual cortex, accurate reconstruction of image stimuli from human fMRI is a still challenging work. In this paper, the capsule network (CapsNet) architecture based visual reconstruction (CNAVR) method is developed to reconstruct image stimuli. The capsule means containing a group of neurons to perform the better organization of feature structure and representation, inspired by the structure of cortical mini column including several hundred neurons in primates. The high-level capsule features in the CapsNet includes diverse features of image stimuli such as semantic class, orientation, location and so on. We used these features to bridge between human fMRI and image stimuli. We firstly employed the CapsNet to train the nonlinear mapping from image stimuli to high-level capsule features, and from high-level capsule features to image stimuli again in an end-to-end manner. After estimating the serviceability of each voxel by encoding performance to accomplish the selecting of voxels, we secondly trained the nonlinear mapping from dimension-decreasing fMRI data to high-level capsule features. Finally, we can predict the high-level capsule features with fMRI data, and reconstruct image stimuli with the CapsNet. We evaluated the proposed CNAVR method on the dataset of handwritten digital images, and exceeded about 10% than the accuracy of all existing state-of-the-art methods on the structural similarity index (SSIM).

## 1 Introduction

Human brain decoding [1, 2, 3] plays an important role in brain-machine interfaces, may be extended to help disabled persons in expressing and motioning, and can also help us explore more about the brain mechanism. In these years, functional magnetic resonance imaging (fMRI) has become an effective tool to monitor brain activities and visual decoding based on fMRI data obtained more and more attention. In contrast to visual encoding [4] that predicts the brain activities in response to visual stimuli, the inverse decoding [5, 6] aims to predict the information about visual stimuli through brain activities. In general, classification, identification and reconstruction of image stimuli based on fMRI data are three main means of visual decoding. Some progresses [7] have been achieved, but the most of previous researches focused on either prediction of its category [8, 9] or identification [4] from a candidate set of image stimuli for the unknown image stimulus. The reconstruction of image stimuli is the full-information and most difficult means of decoding, and fewer studies [10] worked on it. The accuracy of visual reconstruction was restricted by several problems: 1) the complex noise during the acquisition of fMRI data; 2) the high dimensionality and limited number of fMRI data; 3) difficulties when imitating the human visual mechanism to develop the computation models.

Current reconstruction methods mainly focused on some simple or small image stimuli. Thirion et al. [11] reconstructed the simple images by rotating Gabor filters in the passive viewing experiment and imagery experiment for the same subject. Miyawaki et al. [12] achieved the reconstruction of simple binary contrast patterns (resolution: $10\times10$) by linearly mapping fMRI data to each pixel of image stimuli. Van Gerven et al. [13] reconstructed handwritten digits 6 and 9 (resolution: $28\times28$) from the fMRI data based on deep belief network [14]. Schoenmakers et al. [15] tried to reconstruct handwritten English letter of 'BRAINS' (resolution: $56\times56$) from the fMRI data using linear gauss model based on sparse learning. Yargholi et al. [16] employed the gauss network to reconstruct handwritten digits 6 and 9. Fujiwara et al. [17] proposed the Bayesian CCA (BCCA) model based on probabilistic extension of canonical correlation analysis (CCA) model [18] that related fMRI data to image stimuli via a set of latent variables. Wang et al. [19] proposed the deep canonically correlated auto encoders (DCCAE) with nonlinear observation models, and reconstruct each view using the learned representations. Recently, some work [20, 21, 22] proved that features of some layers in CNN behaved strong correlation with the brain activities of particular visual cortex, and some researchers started to apply CNN to visual reconstruction. Using the convolution features of the first layer in CNN, Wen et al. [23] implemented the reconstruction of dynamic video frame by frame, and proposed two-stage cascade neural decoding method based on multivariate linear regression and deconvolutional neural network [24]. They first predicted features by multivariate linear regression, then reconstructed images by feeding the estimated features in a pretrained deconvolutional neural network. Du et al. [25] presented a deep generative multi-view model (DGMM) to regard the visual reconstruction as the Bayesian inference of the missing view. In a word, these studies have suggested that DNN especially CNN could help interpret human brain visual information. However, the accurate reconstruction of image stimuli remains to be challenging.

In conclusion, the visual reconstruction aims to design the mapping from fMRI data to image stimuli. Some methods [11, 12] directly tried to learn a linear or nonlinear mapping based on limited number of samples. However, the linear mapping has limited ability to parse the complex function of visual information processing in human visual cortex, and nonlinear mapping is easy to be overfitting based on limited number of samples and has weak generalization. Depending on some specific architectures of feature representation, some methods [18] firstly mapped the fMRI data to the feature representation of corresponding image stimuli, then tried to employ these features to reconstruct the image stimuli through the architecture. CNN architecture with powerful image feature representation [26, 27], seems a good choice. The current state of the art visual reconstruction methods [23] relied on the linear or nonlinear mapping from fMRI data to the features of specific layer in CNN architecture, then tried to reconstruct image stimuli based on information stream of CNN architecture. However, is the path for the perfectly solving the reconstruction of image stimuli?

As we know, there is a fact [28] that the performance on image classification task [29] is much better than on detection [30] and segmentation [31] tasks in computer vision in spite of both employing CNN architecture. CNN architecture composed of hierarchical convolutional and pooling layers, was firstly designed for invariance including translation invariance, rotating invariance, scale invariance and so on. During the forward propagation in CNN architecture, the extracted features are more and more abstract and the local detailed information such as pose and location which is valuable for detection or segmentation is sacrificed, which lead to the difference of performance on classification and detection or segmentation. CNN can be usually designed for specific procedure of information processing in human visual cortex, such as semantic extracting, or decoding of space location, however, CNN cannot extract diverse features of image stimuli, thus incapable of accurate reconstruction that require various sensory information extracting such as semantic class, orientation, location, scale, and so on. The fact that current reconstruction results were not similar with the image stimuli also proves the inference. Invariance and equivariance are two very important perceptions in visual representation. As shown in the figure 1, the invariance is usually designed for the specific task such as the semantic extracting, at the cost of discarding other features that are not correlated with the semantic information. However, the equivariance is designed for many tasks, because it keeps nearly all the features such as location, pose, orientation, and so on. Essentially, CNN architecture performs the invariance by hierarchical subsampling by pooling layer, instead of the equivariance that keeps the quantity of information unchanged while feature representation. The problem makes it difficult to rely on the incomplete information to perform full-information reconstruction of image stimuli.

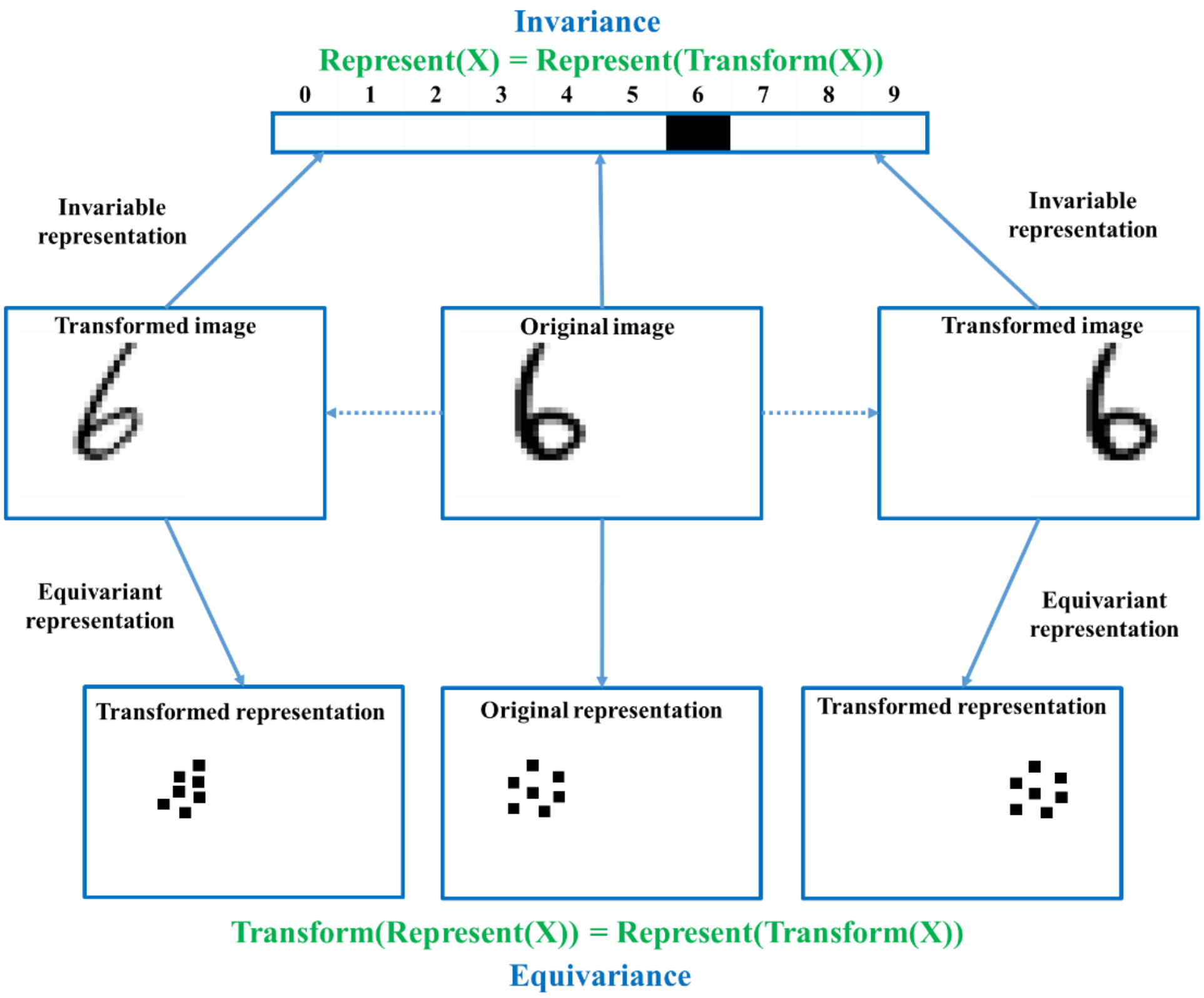

Fig. 1 The difference between invariance and equivariance. Invariance performs the invariable representation when transforming location, pose, and orientation of digit '6', in order to keep invariable class. Equivariance performs the transformed representation when transforming the location, pose, and orientation of digit '6', in order to keep equivariant representation. Invariance is usually designed for the specific task, but equivariance can be used for many tasks, because the equivariant representation contains more diverse features about the image.

In addition, we analyzed it from the perspective of human visual mechanism. After one person glances at one image, he or she can simultaneously answer many questions about the image such as 'what is the main object?', 'where is the object', 'what characteristics does the object have?' and so on. We noticed that these questions contain many characteristics of the image and we can conclude that the procedure of visual information representation in human visual cortex requires the equivariance instead of invariance, because only equivariance can ensure that the semantic class, location, scale, orientation and some other detailed information be preserved instead of discarding, in order to deal with various visual tasks.

According to the above analysis, in order to achieve accurate visual reconstruction, we need an architecture that can keep the equivariance when feature representation and contain diverse features to make it possible to achieve accurate reconstruction from the perspective of information completeness.

Aimed at equivariance instead of simple invariance, Hinton [32] firstly proposed the concept of capsule and designed the promising capsule network (CapsNet) based on convolutional operation and routing by agreement. In CNN architecture, each layer just includes some disorder neurons which makes it hard to perform some organizations of detailed internal structure. However, in the CapsNet, Capsules serve as the basic units of each layer and contain a group of neurons, which can organize some internal structures inspired by the structure of cortical mini column [33] including several hundred neurons in primates. The length of a capsule can predict the presence of a particular object for the invariance, and the features in a capsule can predict the complete attributes of a particular object for the equivariance. The CapsNet reached high accuracy on MNIST [34] digits recognition and demonstrated considerably better results than CNN on highly overlapping digits. Most importantly, the CapsNet provide the equivariance when feature representation. The CapsNet can achieve the accurate classification, which can prove that the capsule features contained the abstract information; can achieve the accurate reconstruction, which prove that the capsule features contained the complete information of image stimuli and is the equivariance of image stimuli.

In this study, we proposed the new CapsNet architecture based visual reconstruction (CNAVR) method that accords well with the human visual information representation in human visual cortex based on the equivariance. We relied on the linear mapping from fMRI data to the high-level capsule features in the CapsNet, then tried to reconstruct image stimuli through CapsNet architecture. To the best of our knowledge, this paper is the first to study visual reconstruction via the new promising capsule network architecture.

## 2 Materials and Methods

### 2.1 Experiment data

The Dataset [16] employed in the study contains a hundred handwritten gray-scale digits (equal number of 6 and 9) at a 28×28 pixel resolution taken from the MNIST database and the fMRI data from V1, V2 and V3. The fMRI data contains 3092 voxels in total. In the experiment, we employed 10-fold cross validation to test the proposed CNAVR method.

### 2.2 Overview of the proposed CNAVR method

In order to achieve accurate visual reconstruction, we introduced the new CapsNet architecture to construct the feature representation of the equivariance between the image stimuli and capsule features, and learned the mapping from the fMRI data to capsule features. Our proposed CNAVR method included two-stage training. Firstly, we employed the CapsNet to train the equivariance from image stimuli to capsule features, and from capsule features to image stimuli in an end-to-end manner by using convolutional, fully connected and routing by agreement operations. After the training, given one input image, we can obtain the corresponding capsule features, and can reconstruct the input image accurately again based on the capsule features, which indicated that the capsule features did not throw away location, pose, scale characteristics and so on for the sake of invariance, and kept complete information of image stimuli when feature representation. Then we selected voxels to reduce dimensionality of fMRI data by encoding performance, and learned the mapping from dimension-decreasing fMRI data to the capsule features using several fully connected layers. After the two-stage training, we realized the mapping from fMRI data to visual reconstruction of image stimuli. Given the fMRI data of one presented image stimulus, we can predict its capsule features about 6 and 9 with the learned mapping, and the accurate reconstruction can be accomplished using the max-length capsule.

### 2.3 Capsule and dynamic routing between capsules

Hinton [32] recently proposed the perception of capsule and dynamic routing between capsules. Each capsule contains a group of neurons. As the equation (1), the capsule $j$ performs the non-linear squashing activation function for the given input vector $\mathbf{s}_j$ in the CapsNet, and output vector $\mathbf{v}_j$. The orientation of vector $\mathbf{s}_j$ is preserved, but the length is squashed between 0 and 1. The parameters in $\mathbf{v}_j$ represent the various properties (position, scale, and texture) of a particular entity, and the length are used to represent the existence of the entity.

$$\mathbf{v}_j = \frac{\|\mathbf{s}_j\|^2}{1+\|\mathbf{s}_j\|^2}\frac{\mathbf{s}_j}{\|\mathbf{s}_j\|} \tag{1}$$

The input $\mathbf{s}_j$ is a weighted sum over all prediction vectors $\hat{\mathbf{u}}_{j|i}$ that is produced by multiplying the output $\mathbf{u}_i$ of a capsule in the layer below by a weight matrix $\mathbf{W}_{ij}$.

$$\begin{aligned} \hat{\mathbf{u}}_{j|i} &= \mathbf{W}_{ij}\mathbf{u}_i \\ \mathbf{s}_j &= \sum_i c_{ij}\hat{\mathbf{u}}_{j|i} \end{aligned} \tag{2}$$

The coupling coefficients $c_{ij}$ are determined by the iterative dynamic routing process. The coupling coefficients between capsule $i$ and all the capsules in the layer above are determined by the softmax of $b_{ij}$ indicating the probability that capsule $i$ should be coupled to capsule $j$.

$$b_{ij} = b_{ij} + \hat{\mathbf{u}}_{j|i} v_j$$
$$c_{ij} = \frac{\exp(b_{ij})}{\sum_k \exp(b_{ik})} \tag{3}$$

Where $b_{ij}$ is initially set to zero, then is iteratively refined by measuring the agreement between the output $\mathbf{v}_j$ and the prediction $\hat{\mathbf{u}}_{j|i}$ made by capsule $i$ in the layer below, using the scalar product $\mathbf{v}_j \hat{\mathbf{u}}_{j|i}$. Three looping can obtain the nice coupling coefficients and routing by agreement essentially tried to learn the relation between part and whole.

2.4 Training image feature representation of equivariance

We employed the CapsNet to perform the equivariance between images and corresponding high-level capsule features. As shown in Figure 2, the CapsNet architecture is shallow with only two convolutional layers, dynamic routing layers and several fully connected layers. Given one image (size: 28×28), the first layer (Conv1) performs 256 convolutional (kernels: 9 ×9) operations with a stride of 1 and ReLU [35] activation. This layer converts pixel intensities to the local features (size: 20 × 20) that are then used as inputs to the primary capsules. The second layer (PrimaryCaps) also performs 256 (32 ×8) convolutional (kernels: 256 ×9 ×9) operations with a stride of 2 to produce 32 capsule maps (size: 6 ×6) whose capsule is an 8D vector. This layer is to construct capsules for dynamic routing operation in next layer. The final Layer (DigitCaps) has one 16D capsule per digit class (from 0 to 9) and each of these capsules receives input from all 1152 (32 ×6 ×6) capsules in the layer below.

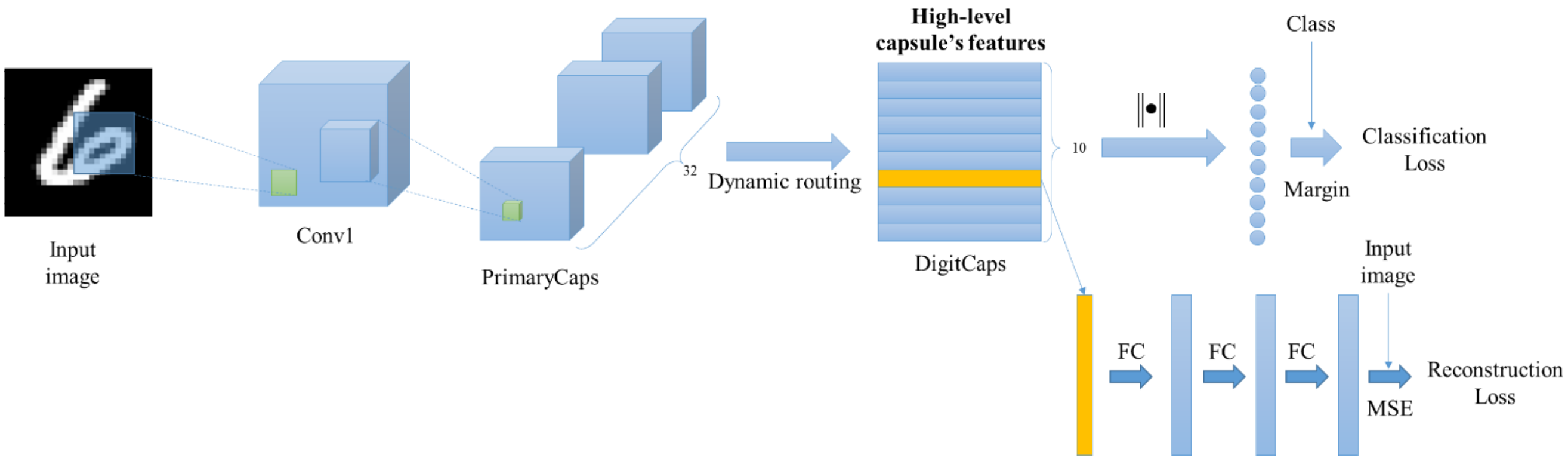

Fig. 2 The architecture of CapsNet. The first two layers perform convolutional operations to construct the primary capsule structure. Each capsule in the PrimaryCaps layer includes 8D features, and each high-level capsule in the DigitCaps layer includes 16D features that include more diverse features. The architecture employs the dynamic routing to replace the pooling operations to avoid the loss of the valuable information such as orientation, location, scale and other detailed features that are important for the equivariance when forward propagation. The 'yellow' mark represents the longest capsule in the DigitCaps and using it can predict the input image based on three fully connected layers.

Because the length of the capsule represents the probability that a specific entity exists, the CapsNet aims to make the corresponding capsule vector longer if some digit is present in the image and make the other capsule vectors shorter. So, the classification loss is simply the sum of the losses of all digit capsules, and defined as shown below. In the training, set $T_c$=1 if a digit of class c is present and set $m^+ = 0.9$, $m^- = 0.1$ and $\lambda$=0.5.

$$L_{Classification} = \sum_{c} \left( T_c \max(0, m^{+} - \|\mathbf{v}_c\|)^2 + \lambda (1 - T_c) \max(0, \|\mathbf{v}_c\| - m^{-})^2 \right) \tag{4}$$

Most importantly, in order to ensure the equivariance of mapping from images to capsule features, the CapsNet adds a decoding network on top of the capsule network. The decoding network contains 3 fully connected layers using ReLU [35] activation function in the first two layer and sigmoid activation function in the output layer. The reconstruction can be accomplished with the corresponding capsule features. The full valuable diverse information for image reconstruction is preserved in the capsule features by minimizing the mean squared error (MSE) between the reconstructed image and the input image.

So, the overall loss $L_{overall}$ is the classification loss plus the weighted decoding loss. The classification loss is to force the high-level capsules distinct in the length for different digits. The reconstruction loss is to force the network to preserve all the information required to reconstruct the image throughout the CapsNet, and acts a bit like a regularization.

$$L_{overall} = L_{classification} + \mu MSE(I, FC_{decoding}(\mathbf{v}_k)) \tag{5}$$

In the experiment, the Adam optimizing [36] was used to avoid overfitting and fasten the training. It is noted that $\mu$ was set 4.0, batch size 10. We finished the training after about 20 epoch based on the MNIST dataset [34] using Tensorflow [37].

2.5 Training the mapping from the fMRI data to capsule features

After finishing the training of the CapsNet, we obtained the architecture of feature representation for equivariance instead of invariance. Given one image stimulus, we can predict its high-level capsules' features and also can reconstruct the image using the longest 16D capsule features. In order to realize the reconstruction of image stimuli from fMRI data, we need to map the fMRI data to capsule features. However, the fMRI data is usually limited because of the acquiring device, subjects, time and other reasons, and the dataset used in this study only contain 100-pair digits and corresponding fMRI data in total. In addition to the problem about the data, the dimensionality of the fMRI data reaches 3092, a very high number compared to the number of the samples.

Faced with the problem of high dimensionality, we chose to use those voxels that are maximally correlated with the image stimuli and the correlation is measured by the effect of fitting namely encoding performance. So, we firstly employed the simple linear regression to fit each voxel with the capsule features instead of images, because the capsule features are the equivariance of image representation, contain all valuable information and have the shorter dimensionality. In detailed, as shown in the Figure 3, given one image, we can predict its 10 high-level capsules, and selected the longest capsule, then employed the 16D vector and the corresponding 3092D fMRI vector as a training sample. In this way, we can make up 90 training samples and employ linear regression to fit each voxel. In order to measure the encoding performance of each fMRI voxel, the goodness-of-fit between model predictions and measured voxel activities was quantified using the coefficient of determination (R2), which indicates the percentage of variance that is explained by the model. Finally we selected those voxels whose R2 is at the top 100, and reduced the dimensionality of fMRI data.

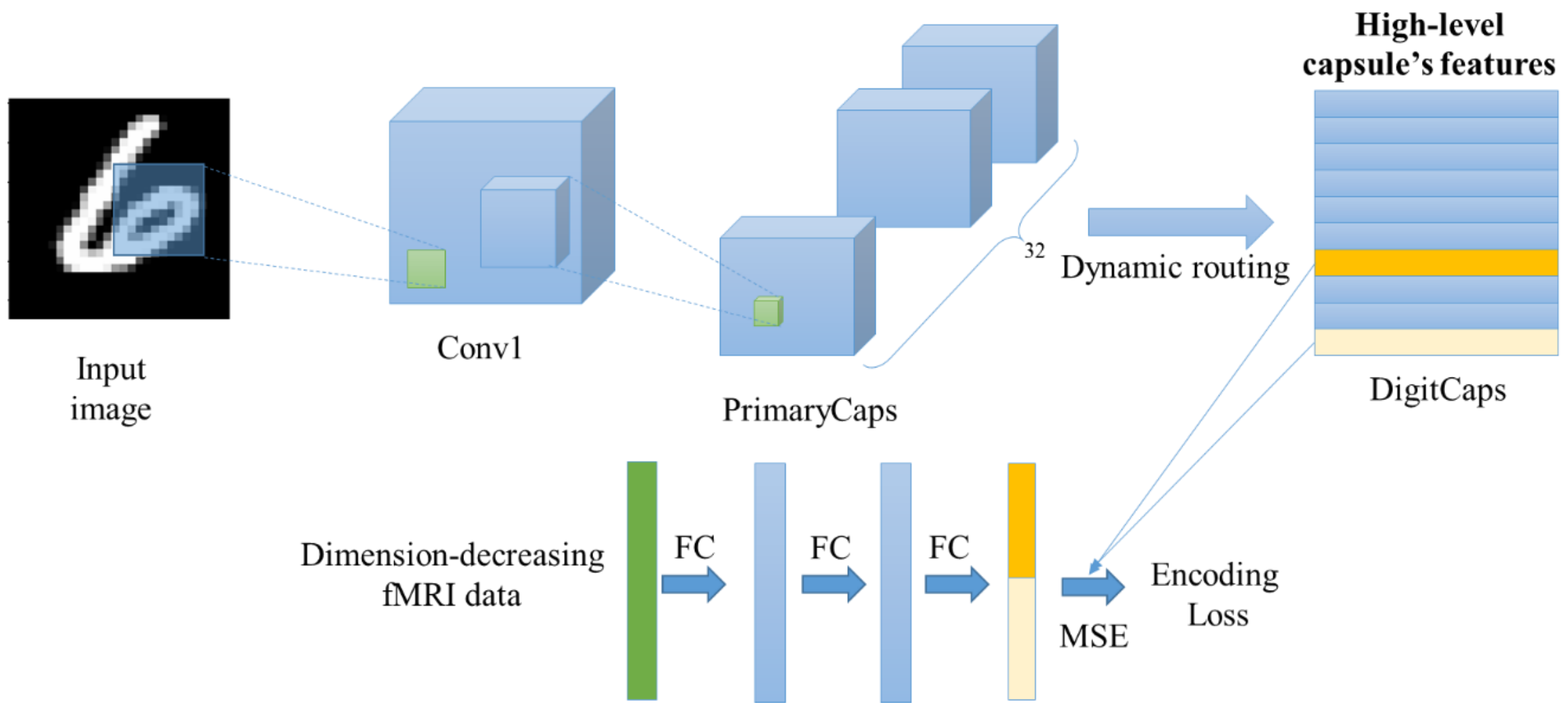

Fig. 3 The architecture from the fMRI data to capsule features. The three fully connected layers are employed to predict 'pale yellow' capsule and 'deep yellow' capsule that represent digit class 6 and 9 respectively. The 'cyan' vector represents the dimension-decreasing fMRI data based on the encoding performance that evaluated by the linear regression of high-level capsule features. The previous trained CapsNet is used to make up training samples (pairs of two capsule features and dimension-decreasing fMRI data) to train the mapping from the fMRI data to capsule features.

Next, we designed the network that maps fMRI data to the two capsules of digits 6 and 9. Our network is composed of three fully connected layer using ReLU activation function in the first two layer and no activation function in the last layer. It should be noted that we try to add the number of layers, but find no benefit in the experiment. The first layer's dimensionality is 256, the second 128, and the last 32. We add the L2 regularization operations in the first two layers to prevent the network from overfitting because the number of training samples is too limited to be easily overfitting. The last layers are split into two 16D vectors in the middle and employed squashing function to resize each length between 0 and 1. Each pair of sample include one input image and one fMRI vector, we first employed trained CapsNet to predict the capsules in response to each input image. The capsules of digits 6 and 9 serve as the ground truth to train the mapping from the 100D fMRI data to it. We employed the mean square error (MSE) to perform gradient descent to update the weight parameters. It should be noted that the weights in the CapsNet is fixed when training the three layers' network.

Similarly, we employed the Adam optimizing method to perform the training. The batch size is set 10, initial learning rate is set 0.0001. The learning curve of the training was shown in the figure below. We finished the training after about 10000 iterations using Tensorflow [37]. From the figure below, we can see that our network behaves well on the 90 training samples and do not suffer from the overfitting because of the regularization and voxels selecting operations. So far, we accomplished the mapping from the fMRI data to capsule features.

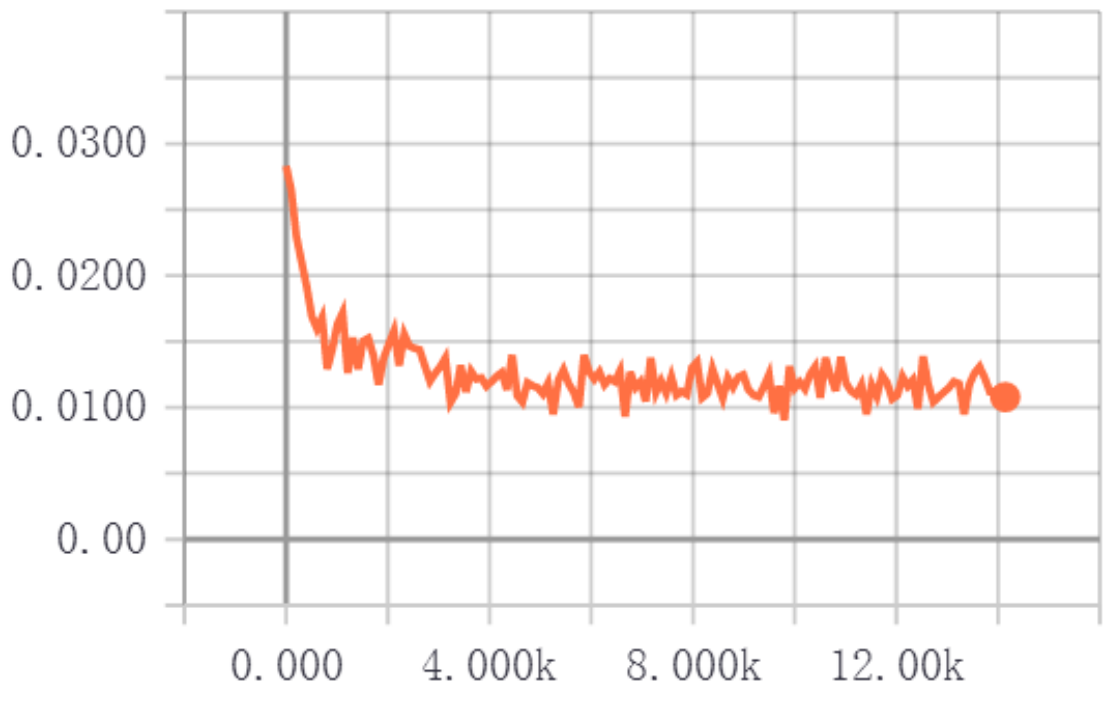

(a) The MSE on training samples

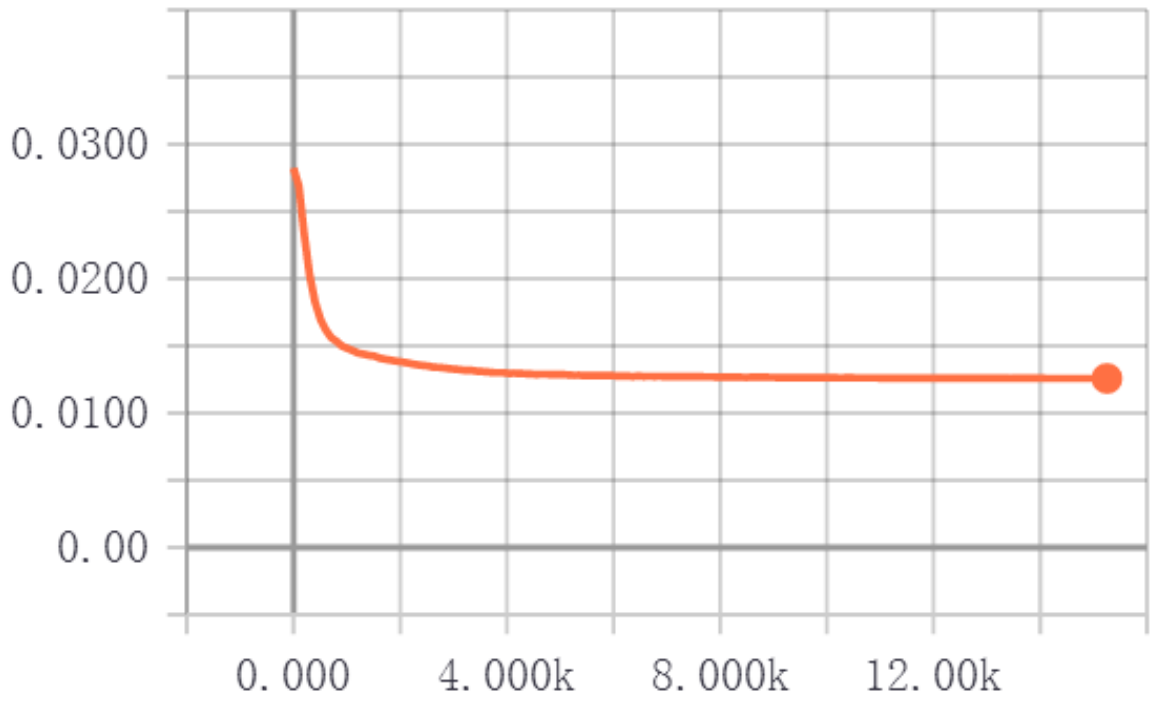

(b) The MSE on testing samples

Fig. 4 The learning curves during training. The mapping from the fMRI data to the capsule features is important for the next reconstruction. The MSE on testing samples is close to that on the training samples, which indicates that our proposed

CNAVR method avoided the overfitting, because of the dimension-decreasing operations and the capsules including the feature information of equivariance. The performance will get improved when obtaining more fMRI data.

### 2.6 Reconstructing image stimuli from human fMRI

After the two-stage training above, we accomplished the equivariance between the images and capsule features, and the mapping from the fMRI data to capsule features. We can reconstruct the image stimuli using fMRI data. As show in the Figure 5, given one fMRI vector, we firstly selected the valuable voxels and predicted its capsule features about the digits 6 and 9 based on the trained three layer' network, secondly we take out the longer capsule, finally we can accomplish the accurate reconstruction by the mapping from capsule features to images in the trained CapsNet.

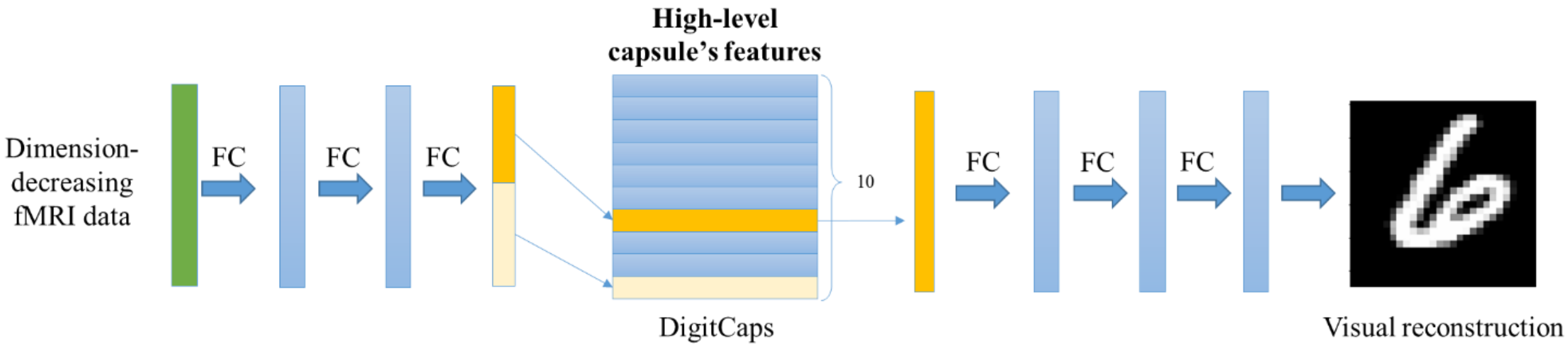

Fig. 5 The flow of visual reconstruction. The former half part trained at the second training stage is responsible for the mapping from dimension-decreasing fMRI data to the two high-level capsules' features, the latter part trained at the first stage is responsible for the reconstruction based on the longer capsule. In this way, the high-level capsule features are used to bridge between the fMRI data and image stimuli.

## 3 Results

### 3.1 The encoding performance

The dimensionality (3092) of the fMRI data is too big to train the mapping from fMRI data to capsule features using limited number (90) of samples, so we selected some valuable voxels according to coefficient of determination R2 reflecting the performance of fitting on training set. As shown in the figure below, we employed the 10-fold cross validation to test our encoding performance for each voxel. We can see that mean correlation coefficient of top-100 voxels reached 0.86 and top-700 voxels exceed more than 0.65. The performance indicated that the capsule features make for better encoding although using the simplest linear regression, which proved the advantage of equivariance of CapsNet when feature representation, In addition, we found that we nearly selected the most of those top voxels using R2, and the selected voxels reached nearly about 85% of top-k voxels in terms of mean correlation coefficient. The comparison demonstrated that selecting voxels by R2 is a good choice and ensure the performance of the next visual reconstruction.

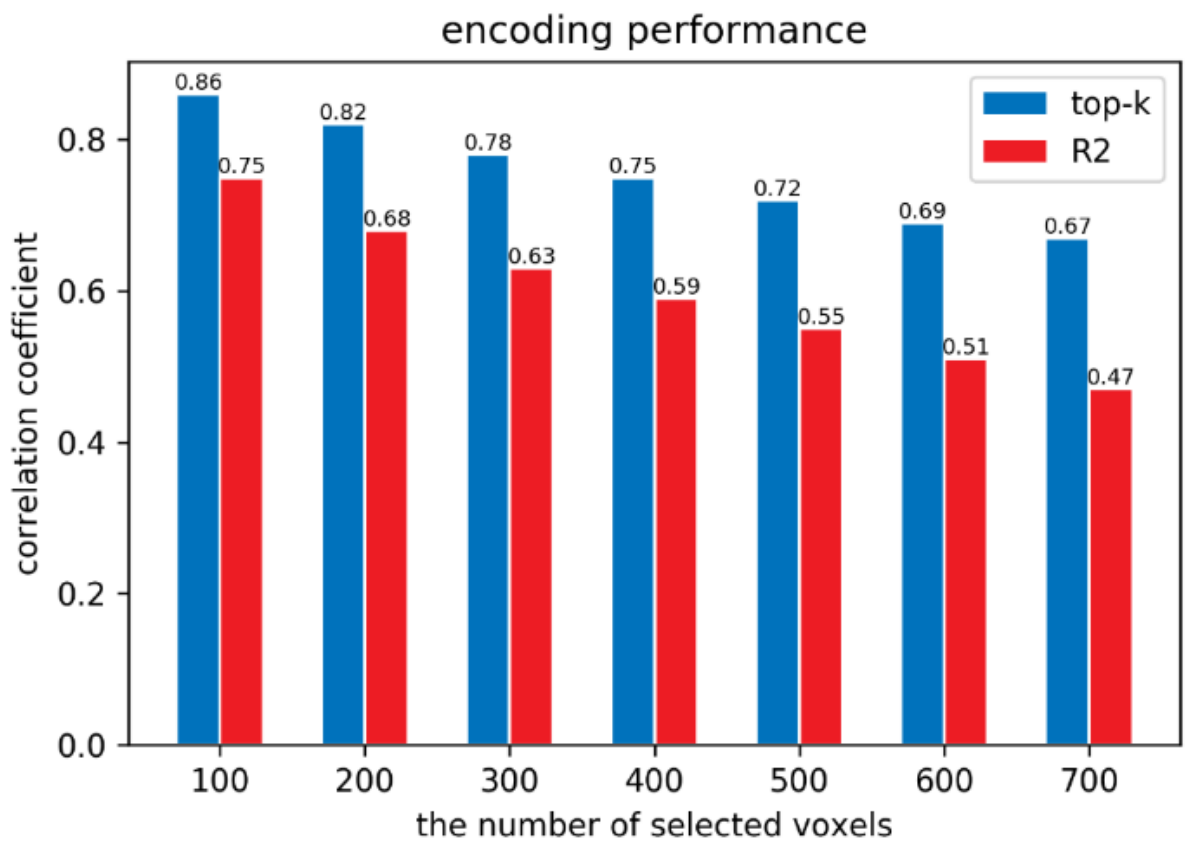

Fig. 6 The encoding performance according to different methods of selecting. The X-axis represents the number of voxels

selected from 3092 voxels. The Y-axis represents the mean correlation coefficient of prediction on testing set for selected voxels by 10-fold cross validation. We can see that employing capsule features can achieve good encoding performance, which indicate that capsules including diverse features such as semantic class, orientation, location, and so on.

3.2 The results of reconstruction

We employed several standard image similarity metrics, including Pearson's correlation coefficient (PCC), mean squared error (MSE) and structural similarity index (SSIM) [38]. Note that MSE and PCC is not highly indicative of similarity, and serves as the auxiliary metrics, while SSIM was proposed to measure structural similarity, can address this shortcoming by taking texture into account and has strong persuasion.

Firstly, we presented the reconstruction results of 12 distinct handwritten digits including the equal number of digits 6 and 9. In order to present the results clearly, we gave the image stimuli, the theoretical reconstruction based on the capsule features of image stimuli, and the visual reconstruction based on the fMRI data. It should be noted that the theoretical reconstruction is obtained based on capsule features using image stimuli throughout CapsNet, and the visual reconstruction are obtained based on predicted capsule features using fMRI data. The theoretical reconstruction demonstrates the theoretical upper limit of our CNAVR method.

From the figure below, we can see that the theoretical reconstruction is perfect and much close to the image stimuli, because the capsule's feature guarantees no missing information when feature representation through trained CapsNet, which proved the equivariance. In addition, our visual reconstruction results are also much similar with the image stimuli, which prove the effective of the proposed CNAVR method. In detailed, we give corresponding quantitative evaluation for each reconstruction in table 1. It cannot be denied that some image reconstruction is not good, as shown in the last column in the figure. We analyzed that the second stage of mapping based on the limited samples remained improvement based on limited fMRI data, and the reconstruction part in the CapsNet is sensitive to the input capsule, and the reconstruction results will change when the capsule features are slightly perturbed. In addition, we can see that these image stimuli indeed do not belong to the common digits. He or she may recall corresponding common pattern if a common subject suddenly looks at the strange kind of image stimuli, which may be an interesting question.

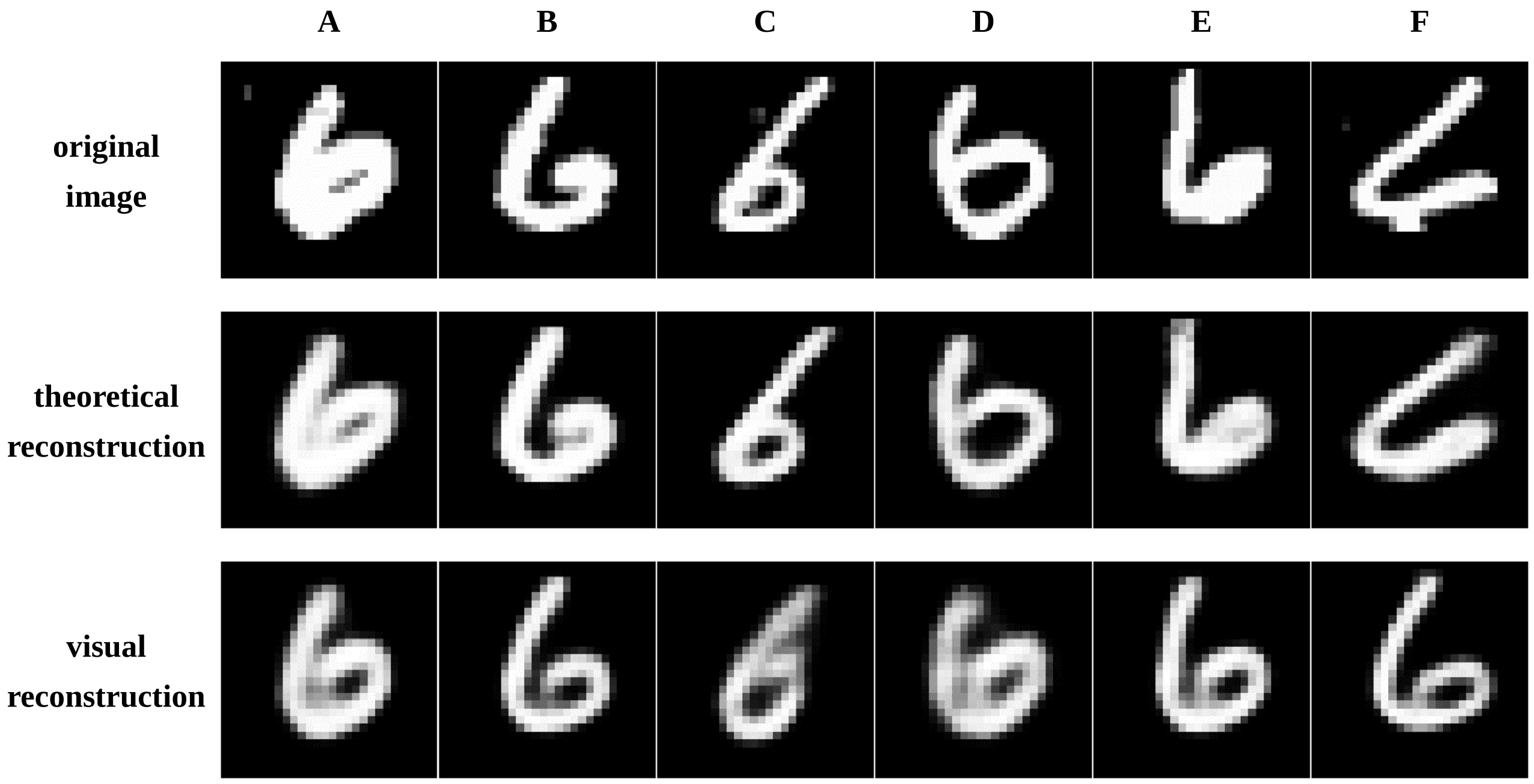

(a) The reconstruction of handwritten digit 6

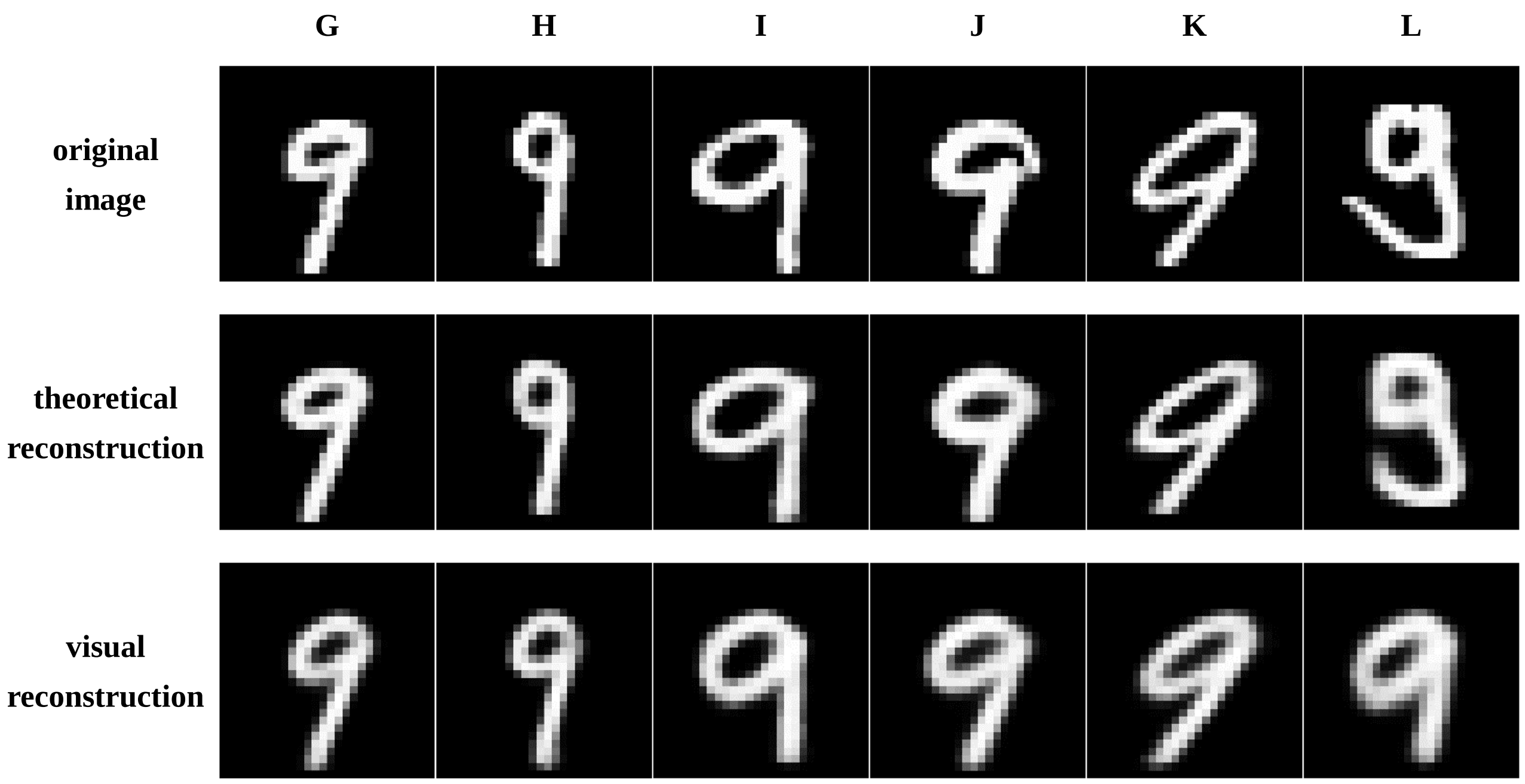

(b) The reconstruction of handwritten digit 9

Fig. 7 Reconstruction results of the proposed CNAVR method. The presented reconstruction includes five high-quality and one low-quality results for (a) and (b). Except for the last column, the reconstruction gets close to the original image stimuli, from the view of orientation, scale, location, and so on. The last column represents the small number of low-quality reconstruction, and the most of low-quality reconstruction belong to the strange image stimuli, which needs further specific improvement.

Table 1 The corresponding quantitative evaluation for each presented reconstruction in Figure 6.

| | **A** | **B** | **C** | **D** | **E** | **F** |
|---|---|---|---|---|---|---|
| **MSE** | 0.023 | 0.021 | 0.029 | 0.037 | 0.048 | 0.090 |
| **PCC** | 0.934 | 0.917 | 0.833 | 0.832 | 0.774 | 0.521 |
| **SSIM** | 0.906 | 0.885 | 0.826 | 0.826 | 0.772 | 0.516 |
| | **G** | **H** | **I** | **J** | **K** | **L** |
| **MSE** | 0.014 | 0.013 | 0.022 | 0.026 | 0.024 | 0.115 |
| **PCC** | 0.912 | 0.900 | 0.890 | 0.873 | 0.869 | 0.460 |
| **SSIM** | 0.901 | 0.898 | 0.888 | 0.867 | 0.866 | 0.459 |

Next, we presented the quantitative results based on 10-fold cross validate compared to the several state of the art methods. As shown in the table 2 below, although the PCC and MSE of our proposed CNAVR method is a little weaker than the current best DGMM [23] and De-CNN [25], the CNAVR exceeds about 10% than them on the most important SSIM metric. We analyzed that the methods should not be crazy about the much high MSE and PCC, because the complex noise in the fMRI data and limited samples reduce the significance of pixel-level comparison. Moreover, human do not care much detailed information in pixel-level and care much of structure according to attention mechanism. So, the two metric just serve as the auxiliary measure, and our proposed methods performed better overall.

Table 2 The quantitative comparison to other state of the art methods.

| Algorithms | MSE | PCC | SSIM |
|---|---|---|---|
| Miyawaki et al. [12] | 0.042 | 0.767 | 0.466 |
| Fujiwara et al. [17] | 0.119 | 0.411 | 0.192 |
| Wang et al. [19] | 0.074 | 0.548 | 0.358 |
| Wen et al. [23] | 0.038 | 0.799 | 0.613 |
| Du et al. [25] | 0.037 | 0.803 | 0.645 |
| **Our CNAVR** | 0.042 | 0.769 | **0.750** |

## 4 Discussion

**The significance of the introduced new CapsNet architecture.** In a regular CNN, there are general several pooling layers. Unfortunately, these subsampling layers tend to lose information for invariance, such as the precise location and pose of the objects. It's really not a big deal if you want to classify the whole image, but it makes it challenging to perform accurate image segmentation, object detection and other tasks which require precise location and pose. Visual reconstruction is exactly the problem that needs to rely on complete characteristic information, which requires the equivariance instead of invariance when feature representation. The CapsNet architecture can be exactly fit for the problem, which benefits from the perception of capsule, dynamic routing and reconstruction regularization loss. In addition, it is obvious that human can simultaneously accomplish many different tasks such as image recognition, image object location, and object pose detection after looking at one image only once. Different tasks always need different characteristics of image and it can conclude that visual information processing in human visual cortex also requires the equivariance. The equivariance ensures that the location, scale, pose and some other detailed information be preserved instead of discarding. The similarity demonstrated that the new CapsNet architecture accords well with the human visual mechanism. Other architecture such as prevailing CNN does not have the equivariance, because they aimed at invariance, abstracting and continues hierarchical abstracting in the process of forward propagation. Our results about encoding during selecting voxels and reconstruction based on the fMRI data both proved the significance of the new CapsNet architecture, which is very promising for visual reconstruction including more complex natural image reconstruction.

**The importance of selecting voxels.** As we know, selecting too many voxels will inevitably introduce more noise and not selecting enough voxels will miss some necessary information, which can both influence the quality of visual reconstruction. So, selecting voxels is an important procedure and challenging problem in visual reconstruction. On one hand, from the mean correlation coefficient of encoding based on top-k and R2 selection, we can find that the means of R2 nearly selected the most of top-k voxels. The second-stage training avoid the overfitting and the 10% higher on SSIM is partly attributed to the performance of selecting voxels, which indirectly indicate the importance of selecting voxels. On the other hand, we employed the much simple linear regression for the encoding to select voxels, and the distance between the theoretical reconstruction and visual reconstruction still remain wide, which indicates that we need to select better voxels to optimize the mapping from the fMRI data to capsule features. How to select the more of top-k voxels and employing nonlinear encoding may be some next choices.

**Two-stage training methods.** We realized the reconstruction with the two-stage training by introducing the new CapsNet architecture that provides the equivariance. We tried to add the third stage of fine tuning that optimize the overall network include the CapsNet and the mapping from fMRI data to capsule features, however, the results did not have the prospective effects. We think that the limited training samples (less than 100) are not suitable for the jointly training. However, the jointly training or end-to-end training is indeed a good direction from the development of computer vision, and the end-to-end training for visual reconstruction based on the CapsNet and more samples may attract more attention in the future.

**Promising reconstruction for complex image stimuli.** We only tested our proposed methods on the public dataset of digits 6 and 9. However, our proposed methods aimed at the reconstruction of digits from 0 to 9. In this study, we mainly aims at how to firstly introduce the new architecture to accomplish visual reconstruction. In addition, the CapsNet is promising for the reconstruction of more complex image stimuli, a more challengeable problem. The reconstruction of more complex image stimuli requires equivariance more, because more complex image stimuli have more complex characteristics and pattern. The next key procedure is to solve the problem of how to better measure the reconstruction results than the MSE error, in order to regularize the CapsNet to preserve the equivariance.

## 5 Conclusion

This paper firstly introduced the new CapsNet architecture for the visual reconstruction, inspired by the equivariance of information processing in human visual cortex. We proposed CNAVR method that provide the equivariance when feature representation. Selecting voxels to reduce the dimensionality of fMRI data and learning the mapping from fMRI data to the capsule feature space are the two key stages in visual reconstruction. We firstly employed the capsule features to estimate the serviceability of each voxel to accomplish the selecting of voxels, secondly we learned the mapping using very few samples. Based on the capsule features, the quality of the two key stages are guaranteed. By qualitative and quantitative results and comparison to the state of the art methods, the CNAVR achieved overwhelming superiority and exceeded by about 10% than the state of the art in the most important SSIM metric. These results demonstrated that our proposed architecture better accords well with the human visual cortex. To the best of our knowledge, this paper is the first to study visual image reconstruction via promising capsule network, which easily lead to recall the breakthrough in the ImageNet classification 2012 [29] competition (the land mark of deep leaning era) by introducing CNN that created about 10% higher classification accuracy compared to traditional methods. Next, in order to achieve better visual reconstruction especially for complex images or videos, the exploration of CapsNet may spring up. There is no doubt that it’s still a start, but a promising start.

## 6 Conflict of Interest

The authors declare no conflict of interest.

## 7 Acknowledgments

This work was supported by the National Key R&D Program of China under grant 2017YFB1002502 and National Natural Science Foundation of China (No. 61701089, No.61601518 and No. 61372172).